\documentclass[letterpaper, 10 pt, conference]{ieeeconf}  % Comment this line out if you need a4paper

\IEEEoverridecommandlockouts                              % This command is only needed if 
\usepackage{balance}
\usepackage{amsmath}
\usepackage{mathrsfs}
\usepackage{amssymb}

\usepackage{pgfgantt}
\usepackage{graphicx}
\usepackage{xcolor}
\usepackage{subcaption}
\usepackage{todonotes}
\usepackage{multirow}
\title{\LARGE \bf
Unifying Local and Global Multimodal Features for Place Recognition\\
in Aliased and Low-Texture Environments
}

\author{Alberto García-Hernández$^{1,2}$, Riccardo Giubilato$^{1}$, Klaus H.\ Strobl$^{1}$, Javier Civera$^{2}$, and Rudolph Triebel$^{1}$% <-this % stops a space
\thanks{This work was supported by the Helmholtz Association, project ARCHES (contract no. ZT-0033), the Spanish Government (projects PID2021-127685NB-I00 and TED2021-131150B-I00) and the Arag\'on Government (DGA-T45-23R).}% <-this % stops a space
\thanks{$^{1}$Institute of Robotics and Mechatronics, German Aerospace Center (DLR)
        {\tt\small <forename>.<surname>@dlr.de}}%
\thanks{$^{2}$I3A, Universidad de Zaragoza, Spain
        {\tt\small jcivera@unizar.es}}%
}

\begin{document}

\maketitle
\thispagestyle{empty}
\pagestyle{empty}

%%%%%%%%%%%%%%%%%%%%%%%%%%%%%%%%%%%%%%%%%%%%%%%%%%%%%%%%%%%%%%%%%%%%%%%%%%%%%%%%
\begin{abstract}
%Aliased and Low-Texture environments are one of the most challenging cases for place recognition. Motivated by this particular application setup, in this paper we propose a novel deep model that 1) leverages multi-modality by cross-attention blocks between vision and LiDAR features, and 2) uses global and local multimodal features. Our experiments in the Mt. Etna dataset, presenting aliasing and low texture, show that our method outperforms significantly the state of the art. We also present results in the less challenging RobotCar dataset, in which our improvement is also smaller.

Perceptual aliasing and weak textures pose significant challenges to the task of place recognition, hindering the performance of Simultaneous Localization and Mapping (SLAM) systems. This paper presents a novel model, called UMF (standing for Unifying Local and Global Multimodal Features) that 1) leverages multi-modality by cross-attention blocks between vision and LiDAR features, and 2) includes a re-ranking stage that re-orders based on local feature matching the top-k candidates retrieved using a global representation. Our experiments, particularly on sequences captured on a planetary-analogous environment, show that UMF outperforms significantly previous baselines in those challenging aliased environments. Since our work aims to enhance the reliability of SLAM in \emph{all} situations, we also explore its performance on the widely used RobotCar dataset, for broader applicability. Code and models are available at {\tt https://github.com/DLR-RM/UMF}.

%[the contribution is unifying global and local multimodal features, and that this is particularly relevant in aliased and low-texture environments like Mt. Etna dataset. There are results with significant improvement in Mt. Etna. And in addition to these results, there are results on RobotCar in which the improvement is not so noticeable, there are ablations, analysis, some qualitative illustrations of results, computational evaluations... ]

\end{abstract}

%%%%%%%%%%%%%%%%%%%%%%%%%%%%%%%%%%%%%%%%%%%%%%%%%%%%%%%%%%%%%%%%%%%%%%%%%%%%%%%%
\section{Introduction}

Simultaneous Localization and Mapping (SLAM) has emerged as a central technology in a multitude of industries including autonomous driving~\cite{bresson2017, burki2019}, automated construction~\cite{mascaro2021}, and agriculture~\cite{shu2021, oliveira2021}. Its development and adoption have been accelerated by advancements in sensor technologies, including multi-camera setups, RGB-D sensors, and more recently, 3D Light Detection and Ranging (LiDAR) sensors, facilitating the construction of large-scale, dense, high-resolution, and consistent 3D maps.%, especially in urban and artificial environments.

% \begin{figure}[!h]
% \centering
% \begin{subfigure}{4cm}
% \centering\includegraphics[width=4cm]{Images/common.png}
% \caption{ Common scenes.}
% \end{subfigure}
% \begin{subfigure}{9cm}
% \centering\includegraphics[width=9cm]{Images/aliased.png}
% \caption{Image and LiDAR aligned.}
% \end{subfigure}
% \caption{ (a) Common scenes from Cambridge landmarks~\cite{kendall2015posenet} and 7-Scenes~\cite{shotton2013scene} datasets, where discriminate features can be extracted. (b) Challenging situations from underwater Aqualoc dataset~\cite{ferrera2019aqualoc} and Mars-Analogue dataset~\cite{meyer2021madmax}.
% Data are severely corrupted with ambiguous elements and low image quality, image extracted from~\cite{zheng20226d}.}
% \label{fig:examples}
% \end{figure}

Many SLAM benchmarks, such as KITTI~\cite{gelfand2003}, Oxford RobotCar~\cite{geiger2013}, KAIST~\cite{choi2018}, and 4Season~\cite{wenzel2020}, are focused on autonomous driving in urban environments. While these datasets incorporate substantial challenges, %such as dynamic objects or large seasonal variations, 
the highly structured scenarios and vehicle-centric perspectives may oversimplify odometry and place recognition tasks.
Other datasets like TUM RGB-D~\cite{sturm2012} and TUM VI~\cite{schubert2018} offer sequences captured using hand-held stereo or RGB-D cameras primarily indoors. However, these sequences are typically short and re-visit previous locations from similar viewpoints and with limited variations in visual appearance. To address these limitations, synthetic datasets such as ICL-NUIM~\cite{schops2019} and TartanAir~\cite{handa2014} simulate more general and challenging motions and environments. However, in the quest for more diversity, {\em unstructured} natural environments still pose the most challenging conditions for visual or LiDAR-based SLAM.

%Over the past few decades, computer vision has evolved into a critical interdisciplinary research area, striving to emulate various aspects of human visual systems. However, even with significant progress in many areas, fully replicating the capabilities of the human visual system remains a formidable challenge. The complex interplay between our understanding of the human vision system's functionality and the limitations of computational resources significantly contributes to this challenge.

%With the surge of LiDAR technology adoption in recent years, especially in autonomous vehicles and robotics, the generation of precise and high-resolution 3D maps has become crucial. These maps are integral to robust and reliable place recognition and navigation in diverse environments.

%One of the most pressing challenges in computer vision is SLAM. Given the increasing interest in mobile robotics and planetary exploration, resolving the SLAM problem is essential for accurately localizing a camera within an unknown environment while simultaneously mapping it. Moreover, the fusion of multimodal data, such as visual and LiDAR information, shows great potential in enhancing SLAM performance, especially in aliased or planetary-like environments.

\begin{figure}[t]
\centering
\begin{subfigure}[b]{0.49\linewidth}
\centering\includegraphics[width=\linewidth]{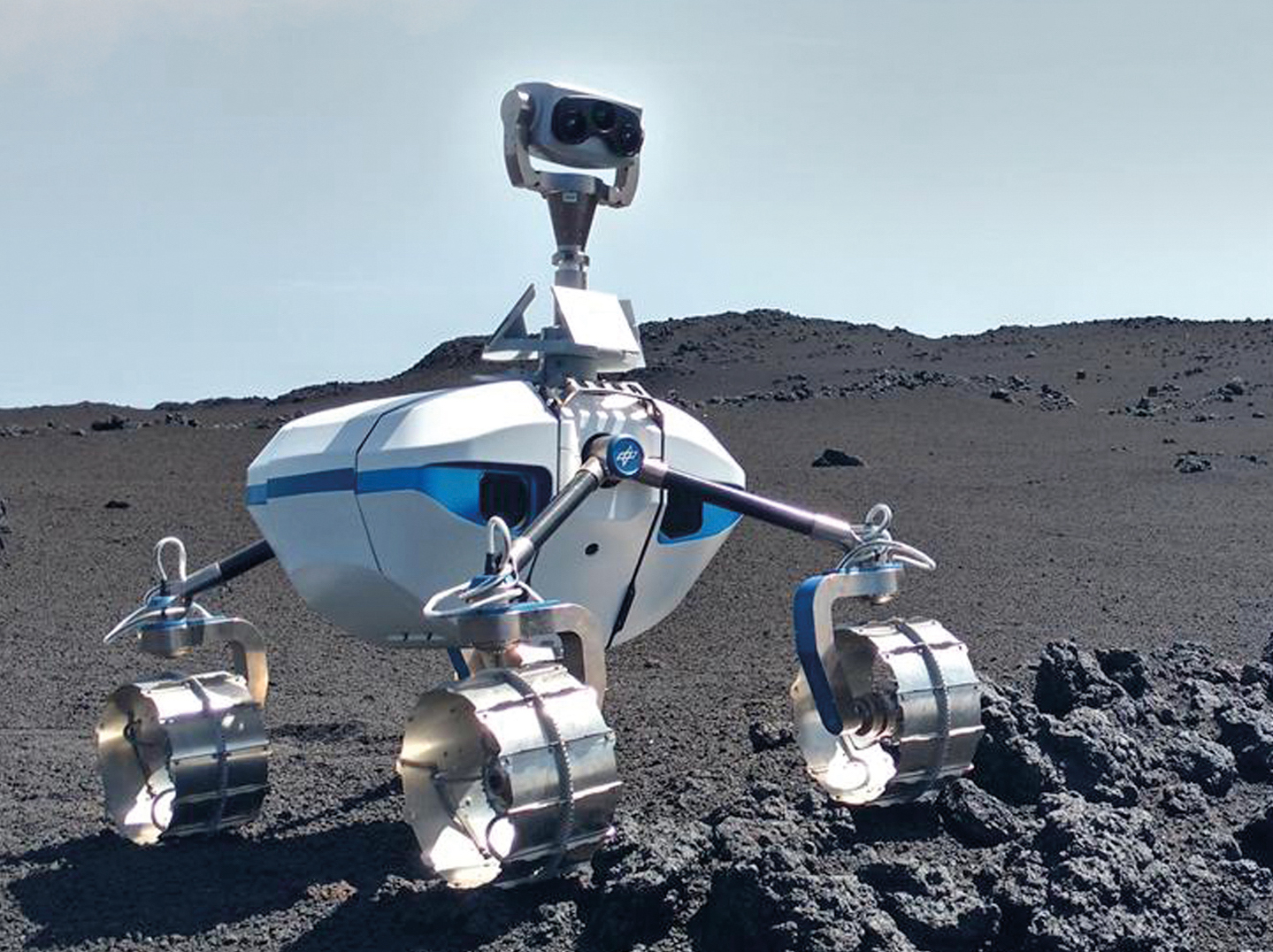}
\caption{The LRU rover traversing a planetary-like environment.}
\end{subfigure}
\begin{subfigure}[b]{0.49\linewidth}
\centering\includegraphics[width=\linewidth, trim=6cm 0 6cm 0, clip]{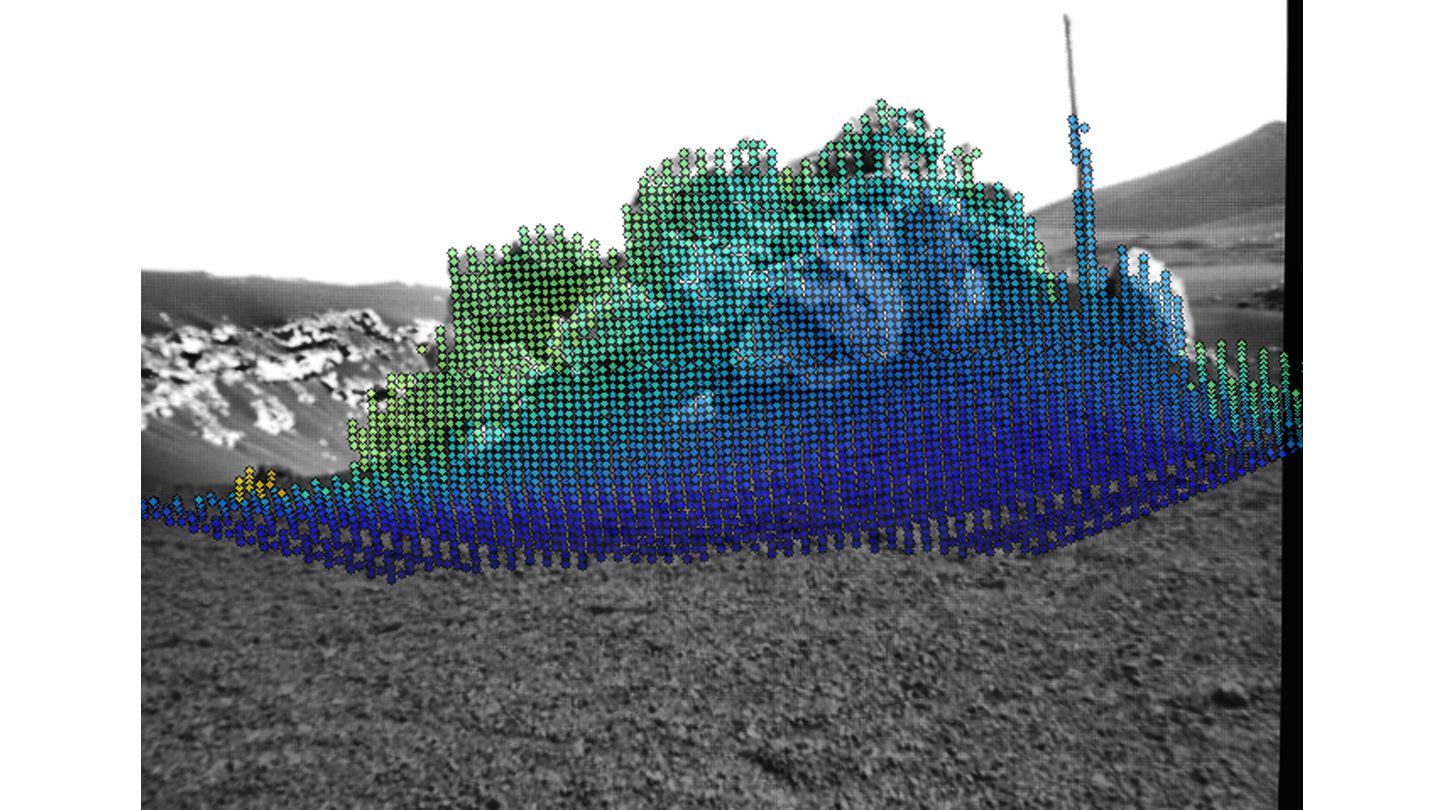}
\caption{Aligned visual and 3D LiDAR data.}
\end{subfigure}

%\begin{subfigure}[b]{0.99\linewidth}
%\centering\includegraphics[width=\linewidth]{Images/rerank2.png}
%\caption{Place recognition pipeline with re-ranking.}
%\end{subfigure}
%\caption{Illustration for the DLR Planetary Stereo Solid-State LiDAR Inertial (S3LI) dataset~\cite{etna} recorded on the Moon-analogue environment of Mt.\ Etna, Sicily.}

\caption{(a) The LRU rover traversing the Moon-analogue environment of Mt.\ Etna, Sicily, recording the DLR Planetary Stereo Solid-State LiDAR Inertial (S3LI) dataset~\cite{etna}. (b) Aligned visual and 3D LiDAR data. Note the challenging texture and geometry for place recognition. %(c) Illustration of the place recognition problem with re-ranking of candidates based on local features (highlighted in red).
}

\label{fig:planetary_env}
\end{figure}

In this paper we propose a novel multimodal place recognition method that we denote as UMF, standing for \underline{\textbf{U}}nifying local and global \underline{\textbf{M}}ultimodal \underline{\textbf{F}}eatures.\footnote{Our naming builds on the title of Ref.\ \cite{unify}.} Our model leverages deep local and global features from visual and LiDAR data, fusing both modalities via cross-attention mechanisms. As our main novelty, we incorporate local feature-based re-ranking to a multimodal setup, showing in our experiments that it leads to a substantial and consistent improvement, in particular in challenging unstructured natural scenes. 

\section{Related Work}

%\subsection{Place Recognition}
%Place recognition is a central task in several areas, including robotics, autonomous driving, and augmented reality. It involves the encoding of sensor data into global descriptors, which are then used to retrieve similar places using appropriate distance metrics in the feature space. The encoding methods are largely dependent on the type of sensor data used. This has led to the development of a wide variety of place recognition approaches, which can broadly be classified into vision-based, LiDAR-based, and visual-LiDAR fusion methodologies. The latter leverages the benefits of both vision and LiDAR data to offer more robust and reliable place recognition, particularly in aliased and low-texture environments.

%\todo{To be vastly summarized, keep related works, avoid explaining base methods}

%Place recognition in robotics and related fields leverages sensor data for feature-based similarity retrieval. Methods are predominantly vision-based, LiDAR-based, and a fusion of both.

%Place recognition techniques have traditionally been divided into vision-based \cite{dbow2, ORBSLAM3}, LiDAR-based \cite{qi2017pointnetplusplus, schubert2018}, and fusion methodologies \cite{MinkLoc, AdaFusion} that leverage both modalities. The latter has shown promise in overcoming the limitations inherent in using either modality alone, particularly in challenging conditions.

%We categorize place recognition techniques by their input data:
Place recognition is a well-established field with a wide literature, relevant applications, and still significant challenges %. The reader is referred to 
\cite{williams2009comparison,garcia2015vision,lowry2015visual,zhang2021visual,masone2021survey,garg2021your,schubert2023visual}.
%for surveys on the topic at different moments in time.
In relation to our research, a first categorization can be done based on the input data modality.
%
%have traditionally been divided into vision-based \cite{dbow2, ORBSLAM3}, LiDAR-based \cite{qi2017pointnetplusplus, schubert2018}, and fusion methodologies \cite{MinkLoc, AdaFusion} that leverage both modalities. The latter has shown promise in overcoming the limitations inherent in using either modality alone, particularly in challenging conditions.
%
{\bf Visual place recognition} has largely evolved from traditional hand-crafted features \cite{dbow2, ORBSLAM3} to learned descriptors \cite{NetVLAD, qi2017pointnetplusplus}, recently moving towards attention mechanisms \cite{Attention-based-VPR, SuperFeatures}. 
{\bf LiDAR place recognition} emphasizes the geometric properties of environments, generally utilizing point cloud data. PointNet \cite{qi2016pointnet,qi2017pointnetplusplus} serves as the foundational architecture for subsequent works such as PointNetVLAD \cite{uy2018pointnetvlad} and LPD-Net \cite{schubert2018}. Attention mechanisms have also been used to improve feature specificity \cite{TransLoc3D, OverlapTransformer}. 
{\bf Visual-LiDAR place recognition} aims at overcoming the limitations inherent in using either modality alone, combining their strengths for more reliable recognition. Techniques like feature concatenation and attention are used in \cite{MinkLoc,AdaFusion}, respectively, which are the main baselines in this area.
{\bf Attention} mechanisms and transformers have shown great utility in several ways. They can serve as patch descriptor filters to spotlight important scene elements \cite{VisualAttention, Attention-based-VPR} or as weight maps that modulate the CNN feature maps to generate global features \cite{TransVPR}. This enables models to focus on salient features and ignore irrelevant information.

Patch-NetVLAD \cite{Patch-NetVLAD} and the work of Zheng {\em et al.}\ \cite{zheng20226d} showcase the benefits of \textbf{unifying local and global features} for visual place recognition. The latter targets the problem of visual aliasing in unstructured scenes \cite{meyer2021madmax}, combining SuperPoint and SuperGlue in the context of 6D camera relocalization. In this work, we propose for the first time unifying local and global features for multimodal data.

\textbf{Self-supervised pretraining} refers to an initial training stage across multiple domains that can then be fine-tuned for specific downstream tasks.
% It enables a model to learn useful features from the data without the need for explicit labeling, which can drastically reduce the cost and effort required for manual annotation. This approach is often used to initialize models before fine-tuning them on a smaller, task-specific labeled dataset, thereby enhancing their robustness and data efficiency.
Here, {\bf contrastive learning methods} like  SimCLR \cite{simclr} and BYOL \cite{byol} take on an important role.
Likewise, {\bf generative self-supervised methods} allow to learn the underlying dependencies in data. In particular, the masked autoencoders do so by learning to fill artificially corrupted parts of the input data, which can only be successfully performed by producing rich, useful data representations.
% Recently, a shift towards generative self-supervised methods, such as masked autoencoders, has been observed, showing superior efficacy in a range of applications. These models learn by reconstructing partially obscured inputs, thereby capturing complex structural dependencies. This generative approach implicitly allows the models to form rich, high-quality feature representations of the data.
Unlike contrastive methods, generative methods do not require carefully designed data augmentations or pair constructions, which are especially challenging on point clouds. For visual data, ConvNeXt \cite{ConvNeXtv2} and Spark \cite{Spark} employ 3D sparse convolutions to apply CNNs to point clouds. For LiDAR data, however, the application of self-supervised learning presents unique challenges. The inherent sparsity and lack of order in LiDAR point clouds, along with the necessity to capture complex geometric and spatial features, make conventional generative methods less suitable. One of the few successful works for generative, self-supervised pretraining on large point clouds is Occupancy-MAE \cite{Occupancy-MAE}. It employs voxelization and masking during the training phase, prompting the model to become ``voxel-aware,'' thereby efficiently leveraging the geometrical and spatial redundancies within the point cloud data. Occupancy-MAE has proven effective in downstream tasks, including 3D object detection and semantic segmentation, even with a high masking ratio of up to 70\%. 
In this work, we employ generative self-supervised pretraining methods in the spirit of the masked autoencoder due to their suitability for multiple data modalities and their lack of dependence on intricate 3D data augmentations. In particular, we build from the Occupancy-MAE due to its proven capability to manage large-scale point clouds with a high masking ratio.

\section{Unifying Local and Global\\Multimodal Features (UMF)}
\label{ch:UMF}

%In the domain of place recognition, the integration of diverse modalities 
%is crucial for tackling challenges such as visual aliasing. 
%is beneficial when aliasing, or lack of recognizable features, affect otherwise challenging unique data streams. 
%Our proposed \textit{UMF} model advances in this direction, effectively merging local and global features across visual and LiDAR modalities, all while leveraging attention mechanisms.

\begin{figure*}[!ht]
    \centering
    \includegraphics[width=0.95\textwidth]{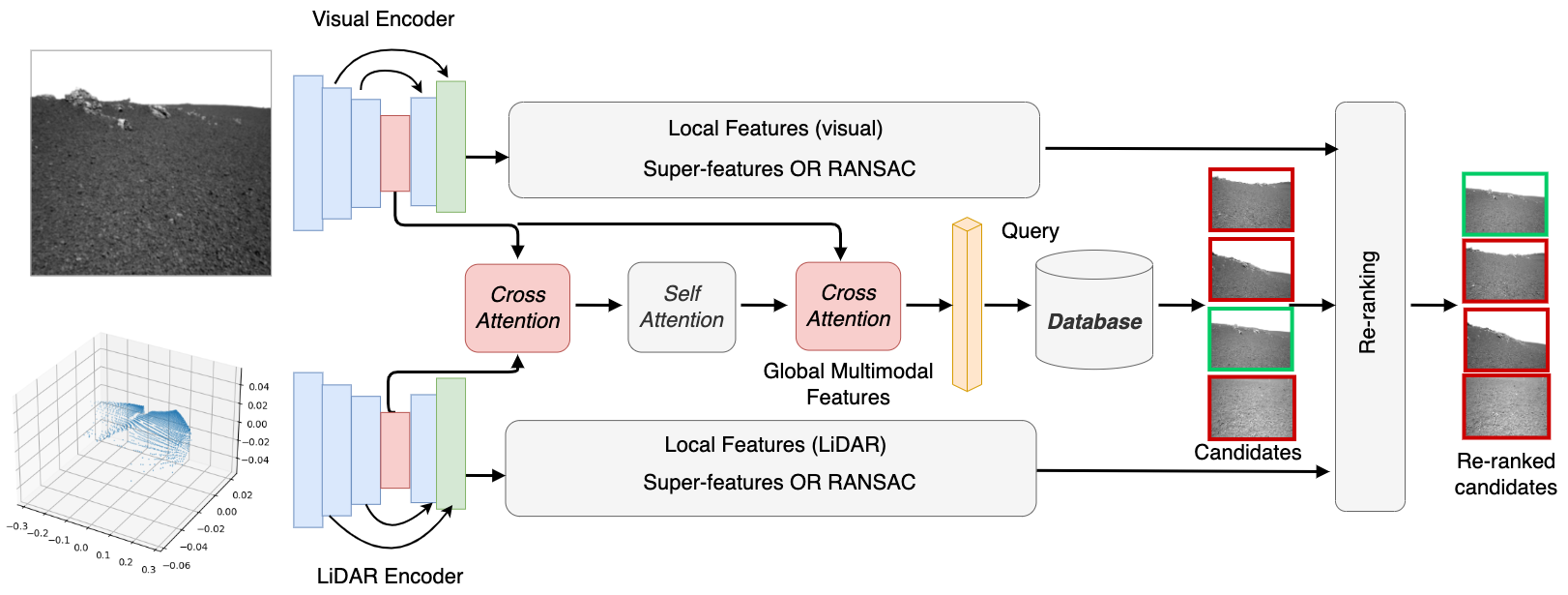}
    \caption{UMF overview. Each branch encodes each of the inputs independently. The encodings of the individual modalities are fused by self- and cross-attention modules into a single global multimodal representation. For each individual data modality, separate branches extract also local features. During inference, we query a database of places with the global multimodal descriptor using a K-Dimensional Tree, the top-$k$ candidates are retrieved via NN-search, and finally they are re-ranked using local features from both modalities. This last stage is the main contribution of our paper.}
    \label{fig:UMF}
    
\label{fig:UMF}
\end{figure*}
%\todo{Use FPN or simple encoder?}
%Our Unifying Local and Global Multimodal Features (UMF) model addresses challenges in aliased and low-texture environments, merging local and global multi-modal features through transformer mechanisms. 
%\todo{It's hard to follow this description in Fig. 2 (add some labels?), and it's hard for this description to be the first paragraph of the Section. Add overall idea first. KS}

%Our model is summarized in Fig.\ \ref{fig:UMF}. It first encodes visual and LiDAR data in two separate representations using two different branches. 
%The merged features form a compact representation that will be used for candidate retrieval. The Feature Pyramid Network (FPN), using a ResNet50 backbone, extracts multi-scale features from visual and LiDAR data. These features are then fused using attention layers.

An overview of our model is showcased in Fig.\ \ref{fig:UMF}. At its core, the UMF design consists of 1) two distinct branches, one for visual data and the other for LiDAR, that encode the data and extract local features, 2) self- and cross-attention blocks to extract a global representation leveraging both modalities, and 3) re-ranking methods that take the top-$k$ candidates by their global representation and re-order them based on local feature matching on each modality. Next, we provide further details on the main aspects of the method.

% Encoding
\textbf{Image and LiDAR Encoding}: Firstly, each of these branches transform the respective input data into low-dimensional representations via a Resnet50 backbone and a LiDAR convolutional encoder~\cite{Occupancy-MAE}. Both encoders follow a Feature Pyramid Network (FPN) architecture to extract multi-scale features, thereby capturing both the local fine-grained details and also relevant patterns with wider extent.

% Attention mechanisms
\textbf{Self- and Cross-Attention}: Following \cite{AdaFusion}, the first work that incorporated attention in multimodal data for place recognition, our UMF model also incorporates attention to enhance its capability to dynamically focus on different parts of the input data.
%UMF utilizes both self-attention and cross-attention mechanisms, as established by \cite{attention} in the original Transformer model.
In the {\bf self-attention} layers, the model assigns different weights of importance to the features within a single modality (either $F_{\mathrm{Vision}}$ or $F_{\mathrm{LiDAR}}$), thereby capturing patterns within local and global contexts. This allows the model to identify distinctive patterns within each modality and enhances its ability to recognize places based on a single modality.
{\bf Cross-attention} layers, on the other hand, take features from both modalities ($F_{\mathrm{Vision}}$ and $F_{\mathrm{LiDAR}}$) as inputs. By interleaving self- and cross-attention layers within our UMF model, it becomes capable of capturing relevant patterns between the two modalities, thereby learning richer scene representations. %Visualizations of attention weights from self attention layers in UMF are shown in sec. \ref{subsec:Qualitative}.

\textbf{Local and Global Features}: 
Building upon \cite{unify}, our UMF model incorporates both local and global features,
%to enhance the representation of the environment, thereby reducing perceptual aliasing. 
%This methodology integrates 
integrating fine-grained details along with the global spatial embedding, and extends the approach to the multi-modal case.
%, refining the model's differentiation capabilities between visually similar locations. 
%The integration of these features builds upon existing methodologies \cite{unify} with further advancements in merging these features for multimodal scenarios. 
We utilize transformers with positional encoding for coarse-level fusion, employing both self- and cross-attention. We implement the same approach introduced in DETR \cite{detr}, which ensures that each element in the feature maps $F_{\mathrm{Vision}}$ and $F_{\mathrm{LiDAR}}$ has a unique positional encoding such that the transformed features exhibit dependencies on relative positions. The relative position dependency enhances the model spatial awareness, as well as inter- and intra-modality relationships between features in the fusion branch.
 Furthermore, such a design consideration promotes rotation and viewpoint invariance, enhancing robustness against perspective changes.

%This enables the model to improve its spatial awareness and the inter- and intra-modality relationships between features in the fusion branch.

%, as ilustrated in \ref{fig:UM}.
%\todo{add DETAILED references to some plot, or no reference :-).}

As the main novelty of our work, we incorporate re-ranking strategies to multimodal place recognition models. Specifically, we evaluate two strategies for re-ranking based on matching local features, the first one using the so-called Super-features \cite{SuperFeatures} and the second one implementing RANSAC geometric verification.

\begin{figure}[tp]
    \centering
    \includegraphics[width=\linewidth]{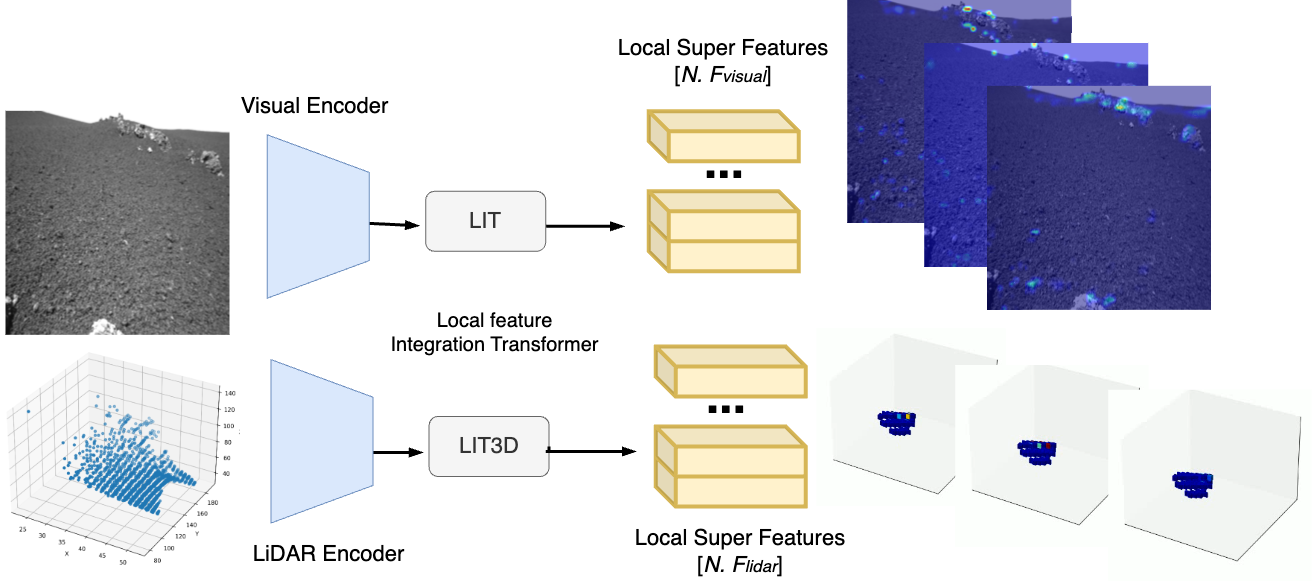}
    \caption{Local Super-features extracted with the LIT module for both modalities. Attention maps show the areas where each Super-feature is focused on.}
    \label{fig:LIT}
\end{figure}

\textbf{Super-Features}: 
%\label{sec:superfeatures}
Super-features~\cite{SuperFeatures} were proposed as mid-level scene representations, showing excellent results in place recognition tasks. The Local Super-features Integration Transformer (LIT) is trained via contrastive learning, passing local features through a transformer layer, as summarized in Fig.\ \ref{fig:LIT}. Specifically, pairs of Super-features are constrained by the contrastive margin loss $\mathcal{L}_{s}$, which minimizes the pairwise distance between matching pairs while simultaneously reducing the spatial redundancy of Super-features $s$ within an image:
\begin{equation}
\mathcal{L}_{s}=\sum_{\mathscr{P}}\left[\left\|s-s^{+}\right\|_2^2+\sum_{n}\left[\mu^{\prime}-\|s-n\|_2^2\right]^{+}\right],
\end{equation}

\noindent where \( \mu \) is a margin hyper-parameter. The index \( n \) corresponds to the Super-features extracted from all negative images in the training tuple, which are compared against a specific Super-feature \( s \). More explicitly, \( n \) serves as a collection of negative samples with the same Super-feature ID as that of \( s \) (denoted as \( i(s) \)).

This process results in a $N \!\times\! F$ ordered set of $N$ Super-features of $F$ dimensions. The construction of Super-features involves an iterative attention module, generating a set where each element focuses on a localized and discriminative image pattern. In this paper, we extend the approach of \cite{SuperFeatures} to 3D scenarios, extracting Super-features from images as well as from voxelized point clouds.

%\todo{which encoder should we use?}

To create Super-features that are complementary, they are encouraged to attend to different ($i \!\neq\! j$) local features, i.e., to different image locations. To do that, the cosine similarity between the attention maps of all Super-features of every image is minimized. Let matrix $
\boldsymbol{\alpha}=\left[\tilde{\boldsymbol{\alpha}}_1, \ldots, \tilde{\boldsymbol{\alpha}}_N\right]
$ denote the $N$ attention maps after the last iteration of LIT. The attention decorrelation loss is then given by:
\begin{equation}
\mathcal{L}_{\mathrm{attn}}(\boldsymbol{x})=\frac{1}{N(N-1)} \sum_{i \neq j} \frac{\tilde{\boldsymbol{\alpha}}_i^{\top} \cdot \tilde{\boldsymbol{\alpha}}_j}{\left\|\tilde{\boldsymbol{\alpha}}_i\right\|_2\left\|\tilde{\boldsymbol{\alpha}}_j\right\|_2} \quad .
%, \quad i, j \in\{1, \ldots, N\},
\end{equation}
%In other words, this loss minimizes the off-diagonal elements of the $N ˆN$ self-correlation matrix of $\alpha$. We ablate the benefit of this loss and others components presented in this section in Section \ref{sec:Ablations}

%\subsubsection{Re-ranking}
%\label{subsec:reranking}
In our UMF model, we implement a re-ranking mechanism to leverage correspondences at Super-feature level. 
%One of the key challenges in our task is the determination of correspondences at the Super-feature level, especially considering that we only have access to pairs of matching images, i.e., image-level labels. To address this, we propose a simple yet effective re-ranking mechanism that relies on nearest-neighbor-based constraints.
As mentioned above, for any Super-feature $s \!\in\! \mathcal{S}$ we have a function $i(s)$ that returns the Super-feature ID, i.e., $i\left(s_i\right) \!=\! i, \forall s_i \!\in\! \mathcal{S}$. Also, let $n(s, \delta) \!=\! \arg \min _{s_i \in \delta}\left|s-s_i\right|_2$ be the nearest neighbor of $s$ from the set $\delta$.

Given a positive pair of images $\boldsymbol{x}, \boldsymbol{x}^{+}$, and two Super-features $\{ s \!\in\! \delta,\, s^{\prime} \!\in\! \delta^{\prime} \}$ from their respective Super-feature sets $\{\delta, \delta^{\prime}\}$, we impose the following criteria to consider the Super-feature pair $\{s, s^{\prime}\}$ eligible:

\begin{enumerate}
\item Reciprocal nearest neighbors: $s \!=\! n ( s ^ { \prime } , \mathcal { \delta } )$ and $s ^ { \prime } \!=\! n ( s , \mathcal { S } ^ { \prime } )$.
\item Pass Lowe's first-to-second neighbor ratio test \cite{lowe2004distinctive}: $\left|\boldsymbol{s}-\boldsymbol{s}^{\prime}\right|_2 /\left|\boldsymbol{s}^{\prime}-n\left(\boldsymbol{s}^{\prime}, \delta \backslash{\boldsymbol{s}}\right)\right|_2 \geqslant \tau$.
\item Have the same Super-feature ID: $i(s)=i\left(s^{\prime}\right)$.
\end{enumerate}

%Formally, these conditions can be expressed as:
%\begin{equation}
%\left(\boldsymbol{s}, \boldsymbol{s}^{\prime}\right) \in \mathscr{P} \Longleftrightarrow\left\{\begin{array}{l}
%{ s = n ( s ^ { \prime } , \mathcal { \delta } ) } \\
%{ s ^ { \prime } = n ( s , \mathcal { S } ^ { \prime } ) }
%\end{array}\right\} \quad \text { and } \quad \left\{\begin{array}{l}
%i(s)=i\left(s^{\prime}\right) \\
%\left|\boldsymbol{s}-\boldsymbol{s}^{\prime}\right|_2 /\left|\boldsymbol{s}^{\prime}-n\left(\boldsymbol{s}^{\prime}, \delta \backslash{\boldsymbol{s}}\right)\right|_2 \geqslant \tau \end{array}\right\}
%\end{equation}
%where $\mathscr{P}$ is the set of eligible pairs, and $\tau$ is a hyperparameter controlling the Lowe's ratio test. 
We set $\tau=0.8$ after empirical considerations.

%Overall, the Super-feature concept and its usage in reranking offer flexibility and significant performance improvement, whilst still maintaining computational and memory efficiency. 

\begin{figure}[tp]
    \centering
    \includegraphics[width=\linewidth]{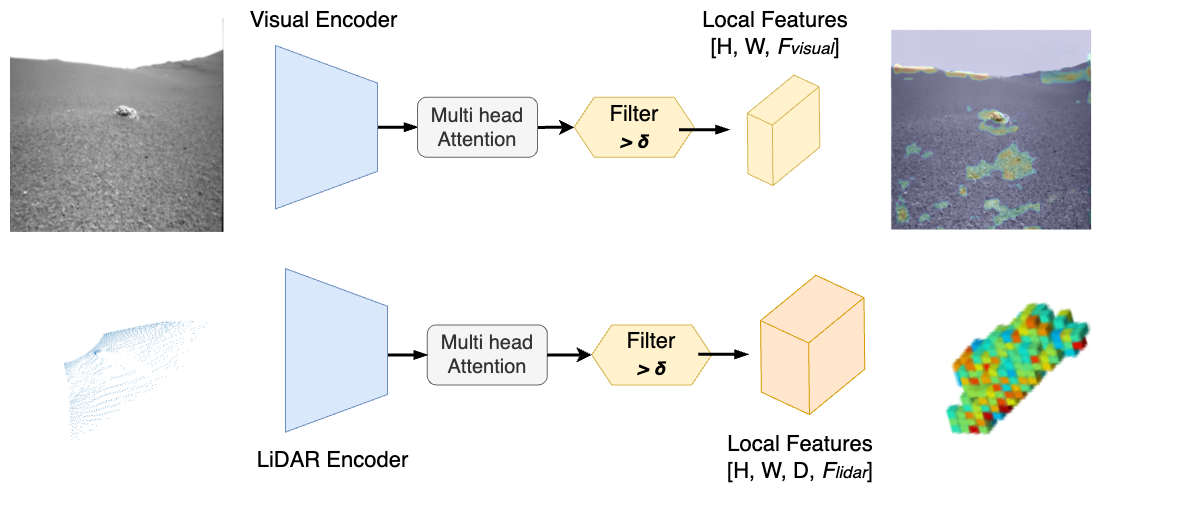}
    \caption{RANSAC local branch visualization. The resulting attention maps are used to select the salient features.}%(reducing potential redundancy and speeding up reranking.}
    \label{fig:ransac-module}
\end{figure}

\textbf{RANSAC}: 
%\label{sec:ransac}
The Random Sample Consensus (RANSAC) variant of our UMF model emphasizes salient feature selection and geometric verification. Initially, a transformer layer is applied to generate attention maps. These maps then filter and pinpoint salient local features%filtering using a hyper parameter $\delta$
, retaining those that carry significant information for place recognition, as shown in Fig.\ \ref{fig:ransac-module}.
The local features are further processed using stacked layers of transformer blocks and we use the attention maps to filter as salient those features for which $\mathcal{L}_{\mathrm{attn}} \!>\! \delta$, $\delta$ being a hyperparameter that we tuned experimentally for both modalitites. Similarly to the Super-features, we project the local features to a 1-D embedding and use a constrastive loss on each local modality during training.

%\subsubsection{Reranking}
%\label{subsec:reranking-ransac}
The RANSAC algorithm is then employed to estimate the geometric transformation between the current observation and candidate locations. This provides a robust approach to matching local features, accommodating the presence of noise and outliers effectively. 
The spatial consistency score is given by the number of inliers returned when fitting a homography between the two images or computing the rigid transformation between voxel grids, using corresponding keypoints computed by nearest neighbor matching.
%While this approach has the potential to deliver higher accuracy owing to the geometric consistency between matched features, it can also be computationally expensive due to the iterative nature of the RANSAC.

\subsection{Training Pipeline}
\label{subsec:training-pretrain}
UMF leverages unlabeled data from similar domains, such as the Mars-analogue in Morocco \cite{meyer2021madmax}, for pretraining. This self-supervised learning approach makes the encoder robust to environmental variations, minimizing the dependency on labeled data and accelerating convergence during fine-tuning for the downstream task.
%Our UMF model takes advantage of self-supervised pretraining methodologies as discussed in section \ref{sec:self_supervised}. Each part involves the use of different pretraining strategies tailored to the unique characteristics of the data.
In detail, it leverages the masked autoencoders for both visual and LiDAR modalities, inspired by Spark \cite{Spark} and Occupancy-MAE \cite{Occupancy-MAE}.
%The main objective of the self-supervised pretraining is to generate robust and informative representations of the input data, which are then transferred to downstream tasks.

\begin{figure}[t]
    \centering
    \includegraphics[width=0.99\linewidth]{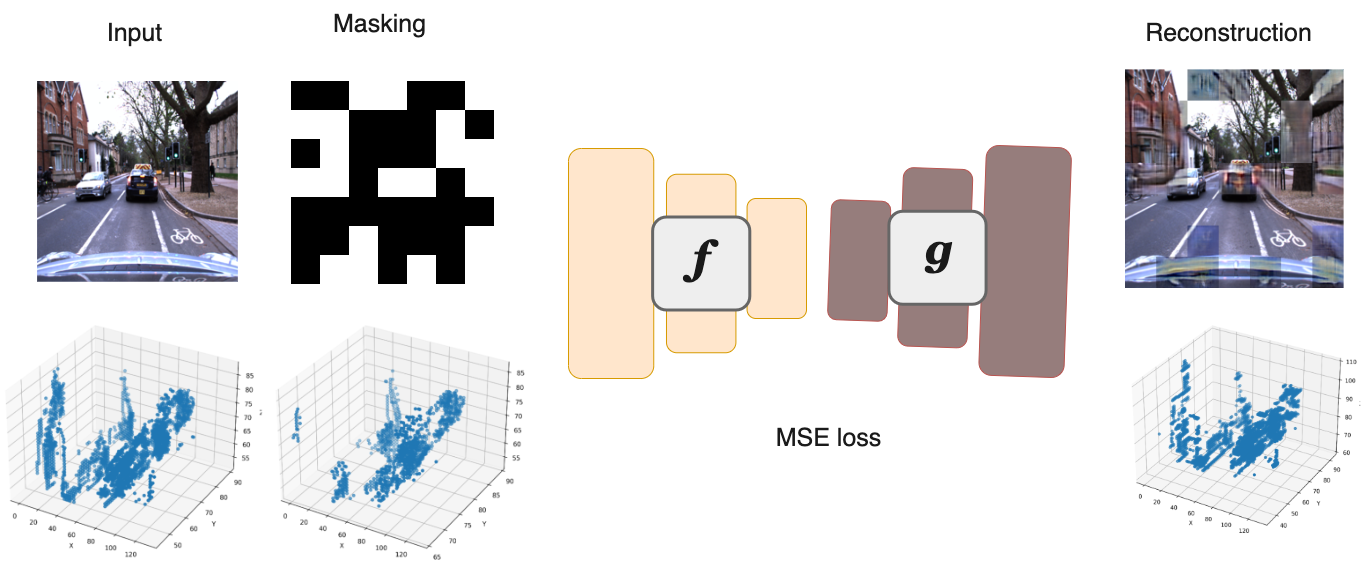}
    \caption{Illustration of our pre-training on RobotCar. First, masked inputs are encoded by \mbox{\boldmath $f$}, followed by the densification process in the decoder \mbox{\boldmath $g$}. After pre-training, only the encoder \mbox{\boldmath $f$} is used for downstream tasks.}
    \label{fig:mae}
\end{figure}

%\subsubsection{Visual Pretraining}
%\label{subsec:pretrain}
\textbf{Visual self-supervised pretraining} follows a contrastive learning approach on a vast unlabeled dataset. It is designed to discern visually similar yet distinct locations, thereby improving the model's ability to tackle visual aliasing.
A patch-wise masking strategy segments images into non-overlapping square patches, each subject to independent masking according to a predetermined mask ratio. 
%The pretraining begins with the patch-wise masking strategy that is commonly employed in masked image modeling. An image is segmented into multiple non-overlapping square patches, each subjected to independent masking according to a predetermined mask ratio. 
%The main challenge lies in obscuring the pixel information from these masked patches without disturbing the data distribution of pixel values, preventing the loss of mask patterns through successive convolution operations, and eliminating unnecessary computations on masked regions.
We follow the approach of \cite{Spark}, where the authors propose to assemble all unmasked patches into a sparse image, which is then encoded using sparse convolutions. 
%This approach ensures no information leakage, maintains compatibility with any convolutional neural network (convnet) without the need for backbone modifications, and effectively manages the issues of "pixel distribution shift" and "mask pattern vanishing." Moreover, sparse convolution only computes at visible places, leading to a more efficient process. When fine-tuning, all sparse convolutional layers naturally transform into ordinary dense ones, as dense images can be considered a specific case of sparse images without "holes".
%The encoding process is hierarchical; the encoder generates a set of feature maps of varying resolutions or scales. For instance, the selcted ResNet model  produces four scales of feature maps, each with different resolutions, which are then used for decoding.
The encoder \mbox{\boldmath $f$} is based on ResNet
% figure from authors of ICLR23 "SparK"?
%\begin{figure}[h!]
%%\centering
%\includegraphics[width=13cm]{Images/sparse_vis_pretrain.png}
%\caption{Sparse masked modeling with hierarchy. To adapt convolution to irregular masked input, visible
%patches are gathered into a sparse image and encoded by sparse convolution. To pre-train a hierarchical encoder,
%we employ a UNet-style architecture to decode multi-scale sparse feature maps, where all empty positions are
%filled with mask embedding. This “densifying” is necessary to reconstruct a dense image. Only the regression
%loss on masked patches will be optimized. After pre-training, only the encoder is used for downstream tasks.}
%\label{fig:sparse_vis_pretrain}
%\end{figure}
and the decoder \mbox{\boldmath $g$} on U-Net, including three blocks 
%$\left[\mathcal{B}_3, \mathcal{B}_2, \mathcal{B}_1\right]$, 
with upsampling layers (see Fig.\ \ref{fig:mae}).
%Prior to the reconstruction of a dense image, it is necessary to densify all the empty positions on sparse feature maps. This process, termed "densifying", involves the use of mask embeddings $\left[\mathrm{M}_4\right]$ to get a dense feature  and projection layers $\phi_4$, in case encoder and decoder have different network widths.
Subsequently, a ``densification'' process, based on mask embeddings, produces dense feature and projection layers. 
%The optimization target is to reconstruct an image, as shown in fig. \ref{fig:pretraining_vis},  from $D_1$ using a head module $h$ that should include two more upsampling layers to reach the original resolution of the input. The authors chose per-patch normalized pixels as targets with an $L^2$-loss and calculated errors only on masked positions. These decisions are based on previous findings indicating that such designs enable models to learn more informative features. Following pretraining, the decoder is discarded, and only the encoder is used for downstream tasks. When fine-tuning, the pre-trained sparse encoder can be directly generalized to dense images without any tuning.

%\begin{figure}[h!]
%%\centering
%\includegraphics[width=16cm]{Images/mae-examples.png}
%\caption{Visualization of the self-supervised pretraining phase for the Robotcar dataset, masking ratio $0.6\%$.}
%\label{fig:pretraining_vis}
%\end{figure}

%The derived representations exhibit robustness against out-of-distribution data and surpass the classification accuracy of fully supervised counterparts on diverse labels. Importantly, the loss computation is restricted to masked patches, thereby circumventing self-reconstruction that may dominate the learning process and obstruct knowledge assimilation.

%\subsubsection{LiDAR Pretraining}
\textbf{LiDAR self-supervised pretraining}
adopts the self-supervised pre-training scheme based on masked autoencoders on voxelized clouds presented in Ref.\ \cite{Hess_2023}.
%
%In contrast to other masked autoencoding works \cite{} that primarily aim to reconstruct the masked parts through a regression task, the pretraining for LiDAR data in this work focuses on predicting the occupancy of the 3D scene. 
%
This encourages the network to reason about high-level semantics to recover the masked occupancy distribution of the 3D scene from a limited number of visible voxels. A binary occupancy classification loss is calculated using cross-entropy:
%between the predicted occupied voxels $\mathbf{P}$ and the ground truth ones $\mathbf{T}$:
%
\begin{equation}
\mathcal{L}_{\mathrm{LiDAR}} = -\frac{1}{n_{\mathrm{batch}}} \sum_{i=1}^{n_{\mathrm{batch}}} \sum_{j=1}^{n_{\mathrm{voxel}}} \mathbf{T}_j^i \log \mathbf{P}_j^i \quad ,
\end{equation}
\noindent where $\mathbf{P}_j^i$ is the predicted occupancy probability of voxel $j$ for the $i$-th training sample and $\mathbf{T}_j^i$ the ground truth probability indicating whether the voxel contains points.

To address the problem of the range-dependent density of LiDAR points, we employ a range-aware random masking strategy \cite{Occupancy-MAE}. This separates the occupied voxels into three groups based on their distance from the sensor. We apply a distinct random masking strategy to each group, with decreasing ratio with increasing distance.

%\subsection{Downstream Task: Place Recognition}
\textbf{Downstream task fine-tuning for place recognition} uses a triplet margin loss and batch-hard negative mining, following the MinkLoc approach \cite{MinkLoc}. The triplets are constructed based on spatial proximity, with a 12 m radius for similarity and over 60 m for dissimilarity. Batch-hard negative mining targets active triplets for effective model refinement. Properly balancing global and modality-specific losses is here of major importance.% and an ongoing research focus.

%Upon completing the self-supervised pretraining phase, we fine-tune the UMF model on place recognition tasks using a triplet margin loss with batch hard negative mining strategy, inspired by MinkLoc \cite{MinkLoc}. Each triplet consists of an anchor, a positive, and a negative example. We define similarity based on spatial proximity, with a radius of 12 meters for similar locations and a distance of more than 60 meters for dissimilar locations, thereby introducing a neutral zone.

%Our training strategy employs batch-hard negative mining to construct informative triplets and disregard less informative ones. Specifically, we focus on active triplets, where the loss exceeds the margin, as they provide valuable insights for model refinement.

%Lastly, we have to balance and adjust the global triplet loss to the individual local modality losses. This balance aids in fostering harmonious interaction between the different branches, which subsequently boosts overall performance. Further research is required to optimally calibrate this interplay, potentially revealing more sophisticated ways of integrating multi-modal data within the UMF framework.

\section{Experimental Results}
We test the UMF model on two very different datasets: the DLR S3LI Dataset for planetary rover exploration \cite{etna} and the Oxford RobotCar dataset for autonomous driving \cite{barnes2020oxford}.

\subsection{Data Setup}

%\label{sec:RobotCar}
\textbf{The Oxford RobotCar dataset} include diverse driving scenarios across varied weather and lighting conditions. Nonground point clouds are down-sampled to 4096 points, while corresponding RGB images are down-sampled from 1280$\times$960 to 224$\times$224. To enhance data diversity and limit overfitting, we randomly sample from 15 closest RGB images during training, while only one RGB image with the closest timestamp is used during evaluation. Similarity is defined based on their spatial proximity: elements within 10 m are deemed similar, those separated by at least 50 m are dissimilar; and those falling between 10 and 50 m are neutral. The dataset is split into disjoint training (21.7k elements) and test (3k elements) areas based on UTM coordinates, following the evaluation protocol and the train/test split (baseline) introduced in \cite{MinkLoc}.

%\label{sec:etna}
%The DLR Planetary Stereo, Solid-State LiDAR, Inertial Dataset dataset 
\textbf{The DLR S3LI dataset} includes sequences captured with a hand-held sensor setup comprising a solid-state LiDAR and a stereo camera. The planetary environment is affected by extreme visual aliasing and lacks of salient visual or structural features. The dataset was split into training and validation set, using \textit{s3li\_loops} and \textit{s3li\_traverse\_1} for testing, as they contain overlapping areas.

%\label{sec::pretraining}
\textbf{Additional datasets for pre-training and fine-tuning }
were used to increase the model's robustness and performance. For unstructured planetary environments, we incorporate the MADMAX dataset \cite{meyer2021madmax}, the Erfoud dataset \cite{lacroix2020erfoud}, and the
Long Range Navigation Tests (LRNTs) 
% https://www.dlr.de/rm/en/desktopdefault.aspx/tabid-13819/23995_read-58297/
%providing sequences from a hand-held device on the Moroccan desert, as well as sequences captured from a planetary-like rover on a volcanic slope 
\cite{arjun_etna}. To overcome the scarcity of datasets that contain both visual and LiDAR modalities,
%, that represent challenges of localization in extremely unstructured environments, which contain both visual and LiDAR modalities,
we also generated synthetic sequences with the OAISYS photorealistic simulator \cite{Mueller2021}.

\begin{table}[!h]
\centering
\begin{tabular}{|l|c|c|}
\hline
\textbf{Dataset} & \textbf{\# Samples} & \textbf{Modalities used} \\ \hline
MADMAX \cite{meyer2021madmax} & 11,000 & Stereo \\
Erfoud \cite{lacroix2020erfoud} & 7,000 & Stereo, LiDAR \\
LRNTs \cite{arjun_etna} & 5,600 & Stereo \\
OAISYS \cite{Mueller2021} & 10,000 & Stereo, LiDAR, Instance, Semantic \\ \hline
\end{tabular}
\caption{Datasets used for pre-training and fine-tuning.}
\label{tab:additional-datasets}
\end{table}

%By integrating data from these multiple sources, we aim to address the dearth of multimodal datasets suitable for training models that can operate effectively in extremely unstructured environments.

\subsection{Implementation Details}
All models were implemented in PyTorch \cite{PyTorch} and trained on a compute cluster equipped with 8 NVIDIA RTX 3090 GPUs. For fine-tuning, the learning rate was set to $1\mathrm{e}\!-\!5$ and reduced by a factor of $1\mathrm{e}\!-\!1$ upon plateauing. The model was trained for 200 epochs using Adam \cite{kingma2014adam}. The input image size was set to 224$\times$224. When re-ranking global feature retrieval
results with local feature-based matching, the top 25 ranked images from the first stage are considered.

\textbf{Super-features}: 
The score for Super-features was determined using $\mathcal{L}_{\mathrm{attn}}$. The attention maps were generated at image size of 56$\times$56 and voxel size of 50$\times$50$\times$50. Super-features were represented as a tensor of size $[N, F]$, and the dimensions for the visual and point cloud features are set to 128 and 32, respectively. The final ranking is based on the number of matching features that satisfy the criteria described in Sec.\ \ref{ch:UMF}.

\textbf{RANSAC}: 
We use a multi head transformer encoder in our model to process the fine features. The model returns the average of all attention maps and selects an optimal threshold to identify keypoints. The output consists of attention maps for the image and voxel, of sizes $[N, 56,56]$ and $[N,50,50,50]$, respectively, where $N$ is the number of keypoints. The feature maps for the image and voxel are $[56,56,128]$ and $[50,50,50,32]$, respectively. The re-ranking of candidates is based on the total number of inliers, either for one or both modalities: $score = \#\mathrm{inliers}_{\mathrm{Vision}} + \#\mathrm{inliers}_{\mathrm{LiDAR}}$.

\label{subsec:normalized}
\textbf{Similarity Threshold}: The threshold $\theta$ plays a pivotal role in our model's ranking process, acting as a cut-off value to distinguish between similar and dissimilar pairs during re-ranking. Given two samples \( S_1 \) and \( S_2 \), with a derived similarity score \( \mathrm{sim}(S_1, S_2) \), they are classified as recognizing the same place if $\mathrm{sim}(S_1, S_2) \!>\! \theta$, where $\theta$ adjusts the model sensitivity when classifying candidate matches as true or false positives, and it is determined empirically through cross-validation to optimize precision and recall rates.

\subsection{Comparison against Baselines}

\begin{figure}[!b]
    \centering
    \includegraphics[width=\linewidth]{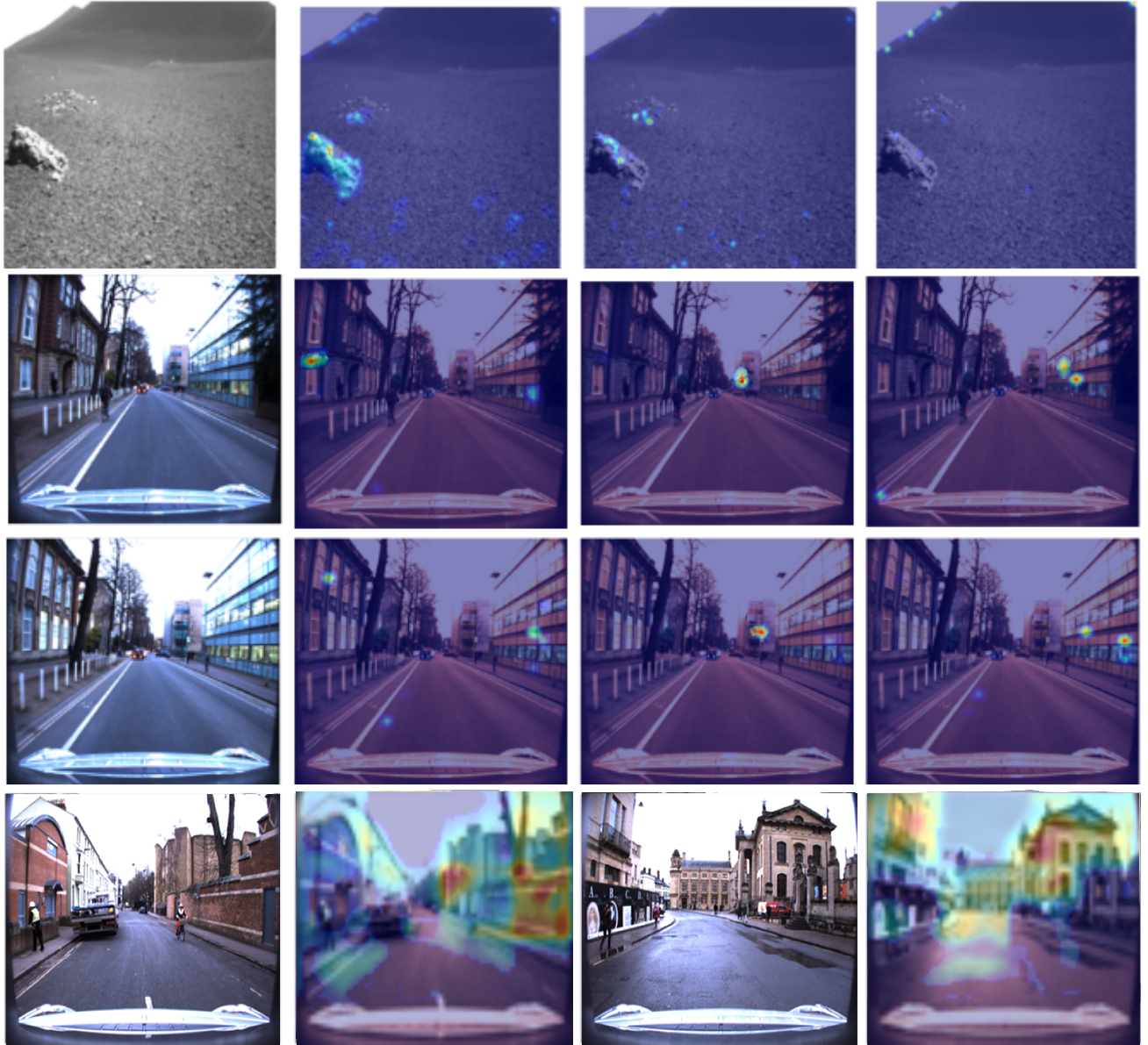}
    \caption{Illustration of the Super-features attention maps generated by the Learning Iterative Transformer (LIT), for S3LI (1st rows) and RobotCar (2nd and 3rd row). The first three Super-features are highlighted, showing the model's propensity to focus on semantic patterns such as rock formations and terrain structures. The last row shows the attention map extracted from the multi head attention for a pair of RobotCar samples.}
    \label{fig:examples}
\end{figure}

Table~\ref{tab:etna-results} shows the comparative performance of our UMF models vs.\ relevant baselines on the S3LI dataset.
%The data is segmented according to modality—Visual, LiDAR, and Multimodal—for clarity and concise evaluation.
Our best model \textbf{outperforms the best baseline by more than 2\% in the three metrics chosen}, which can be attributed mainly to the use of local features for re-ranking.
While the LiDAR data has a limited field of view, they still provide a valuable input under challenging conditions by reducing uncertainty. Moreover, the accurate depths play a crucial role in establishing correct positive pairs.
Also, the geometric verification using RANSAC shows a significant impact in aliased environments, outperforming other approaches. 
%It is noteworthy that it enhances the robustness of the final predictions. 
%Conversely, Superfeatures struggle to consistently focus on the most salient regions when there are few landmarks, which may be a consequence of the decorrelation loss at the attention maps level. These features are compelled to investigate different areas.
%
For an illustrative understanding, Fig.~\ref{fig:examples} depicts the Super-features attention maps generated by our LIT.

\begin{table}[!h]
\centering
\begin{tabular}{|c|c|c|c|}
\hline
\textbf{Method} & \textbf{Recall@1} & \textbf{Recall@5} & \textbf{Top 1\% recall} \\ \hline
\multicolumn{4}{|c|}{\textbf{Visual}} \\ \hline
DBoW2 & 37.44 & 66.1 & 68.12 \\
NetVLAD & 67.2 & 75.5 & 78.3 \\
MinkLoc++ (Vision) & 68.8 & 77.3 & 79.2 \\ \hline
\multicolumn{4}{|c|}{\textbf{LiDAR}} \\ \hline
PointNet++ & 48.41 & 67.77 & 71.8 \\
MinkLoc++ (LiDAR) & 42.4 & 65.8 & 69.4 \\ \hline
\multicolumn{4}{|c|}{\textbf{Multimodal}} \\ \hline
MinkLoc++ & 71.4 & 80.1 & 85.2 \\
AdaFusion & 73.1 & 82.3 & 87.2 \\
UMF (only global feat.)  & 73.5 & 82.9 & 87.5 \\ 
UMF (Super-features) & 75 & 85.1 & 89.1 \\ 
\textbf{UMF (RANSAC)} & \textbf{75.3} & \textbf{85.3} & \textbf{89.5} \\
\hline
\end{tabular}
\caption{Comparison against baselines on S3LI (Mt.\ Etna), for single- and multi-modality.}
\label{tab:etna-results}
\end{table}

The evaluation on the RobotCar dataset, in a very different application domain and with different sensor specifications, demonstrates the robustness and adaptability of UMF.  Table \ref{tab:robotcar-results} shows multimodal results compared against the baselines MinkLoc++ and AdaFusion, in which our UMF model with re-ranking outperforms again both baselines. 

\begin{table}[!h]
\centering
\begin{tabular}{|c|c|c|}
\hline
\textbf{Method} & \textbf{Recall@1}  & \textbf{Top 1\% recall} \\ \hline
\multicolumn{3}{|c|}{\textbf{Multimodal}} \\ \hline

MinkLoc++ & 96.7  & 99.1 \\
AdaFusion & 98.1 &  99.2 \\
UMF (only global feat.) & 97.9 & 99.1  \\ 
UMF (Super-features) & 98.1  & 99.1\\ 
\textbf{UMF (RANSAC)} & \textbf{98.3}  & \textbf{99.3} \\ \hline
\end{tabular}
\caption{Comparison against baselines on RobotCar.}
\label{tab:robotcar-results}
\end{table}

\begin{table*}[!ht]
\centering
\begin{tabular}{|l|l|c|c|c||c|c|c|}
\cline{3-8}
\multicolumn{2}{c|}{} & \multicolumn{3}{|c||}{S3LI (Mt.\ Etna)} & \multicolumn{3}{|c|}{RobotCar}\\ 
\hline
\textbf{Modality} & \textbf{Method} & \textbf{Recall@1} & \textbf{Recall@5} & \textbf{Top 1\% recall} & \textbf{Recall@1} & \textbf{Recall@5} & \textbf{Top 1\% recall} \\ \hline
\multirow{2}{*}{Vision} & UMF (Super-features) & 74.5 & 84.3 & 89 & 98 & 98.4 & 99.1 \\ 
& UMF (RANSAC) & 75.1 & 84.9 & 89.3 & 98.1 & 98.5 & 99.2 \\  \hline
\multirow{2}{*}{LiDAR} & UMF (Super-features) & 73.7 & 83.4 & 87.5 &  97.8. & 98.2 & 99.1 \\ 
& UMF (RANSAC) & 73.9 & 83.8 & 87.8 &  98 & 98.3 & 99.1 \\   \hline
%UMF  & 73.5 & 82.9 & 87.5 \\ 
\multirow{2}{*}{Vision+LiDAR} & UMF (Super-features) & 75 & 85.1 & 89.1 & 98.1 & 98.5 & 99.1 \\ 
& \textbf{UMF (RANSAC)} & \textbf{75.3} & \textbf{85.3} & \textbf{89.5}  & \textbf{98.3} & \textbf{98.8} & \textbf{99.3} \\ \hline
\end{tabular}
\caption{Ablation study on S3LI and RobotCar. Note how the multimodal (Vision+LiDAR) setup presents the best metrics, effectively leveraging both modalities. Observe also how our RANSAC variant consistently outperforms Super-features. Finally, note how the improvement offered by our methods is bigger in S3LI than in the almost saturated RobotCar.}
\label{tab:etna-ablation-results}
\end{table*}

\subsection{Ablation Studies}

%\subsubsection{Re-ranking}
\textbf{Re-ranking}:
To analyze the role of the re-ranking module in our approach, we conduct experiments involving variations of it, such as with and without, for one or all modalities.
The improvements after re-ranking shown in Table \ref{tab:etna-ablation-results} substantiate its importance. On the S3LI dataset, we observe that the visual modality outperforms LiDAR. The re-ranking step on LiDAR data shows marginal improvements. The fusion of both modalities, however, does aid in overcoming visual challenges such as poor lighting or aliasing in both datasets.
Notably, RANSAC emerged as a clear winner over Super-features within the re-ranking methods. This advantage, however, comes at the expense of an increased computational cost.
%

%\begin{table}[!t]
%\centering
%\begin{tabular}{|c|c|c|c|}
%\hline
%\textbf{Method} & \textbf{Recall@1} & \textbf{Recall@5} & \textbf{Top 1\% recall} \\ \hline
%\multicolumn{4}{|c|}{\textbf{Base}} \\ \hline
%UMF  & 97.9 & 98.3 & 99.1 \\  \hline
%\multicolumn{4}{|c|}{\textbf{Visual}} \\ \hline
%UMF (superfeat visual) & 98. & 98.4 & 99.1 \\ 
%UMF (RANSAC visual) & 98.1 & 98.5 & 99.2 \\  \hline
%\multicolumn{4}{|c|}{\textbf{LiDAR}} \\ \hline
%UMF (superfeat pc) &  97.8. & 98.2 & 99.1 \\ 
%UMF (RANSAC pc) &  98. & 98.3 & 99.1\\   \hline
%\multicolumn{4}{|c|}{\textbf{Multimodal}} \\ \hline
%UMF (superfeat all) & 98.1 & 98.5 & 99.1 \\ %\hline
%\textbf{UMF (RANSAC all)} & \textbf{98.3} & \textbf{98.8} & \textbf{99.3} \\ \hline
%\end{tabular}
%\caption{Ablation study on RobotCar. Again, the multimodal setup effectively leverages both data modalities.}
%\label{tab:robotcar-ablation-results}
%\end{table}

%Fig. \ref{fig:rcandidates_etna} contains a quantitative comparison of both ranking approaches and the baseline methods where we study the impact of the number of candidates. The measurement can vary significantly depending on the dataset used, but we found taking the top 20 candidates is a reasonable trade-off for most use cases.

%\begin{figure}[tp]
%%\centering
%\includegraphics[width=\linewidth]{Images/recall_candidates_etna.png}
%\caption{Comparison of top 1 recall of both UMF reranking approaches depending on the number of candidates used in the Etna dataset.}
%\label{fig:rcandidates_etna}
%\end{figure}

\begin{figure}[!t]
\centering
\includegraphics[width=.9\linewidth]{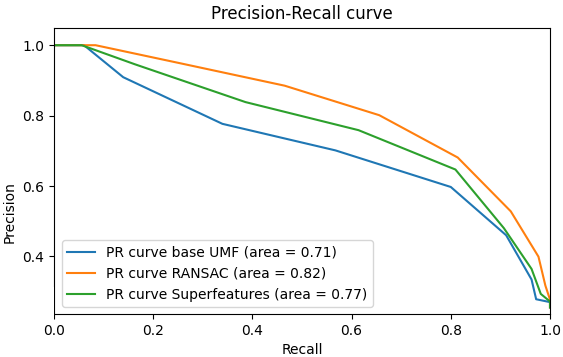}
\caption{Precision-recall curves on S3LI (Mt.\ Etna). We compare the base UMF model with both re-ranking variants.}
\label{fig:precision_recall}
\end{figure}

The normalized similarity threshold \( \theta \) defined in Section \ref{subsec:normalized} has also been adjusted 
%In our final analysis, we adjusted the normalized similarity threshold \( \alpha \), which was initially defined in Section \ref{subsec:normalized},
to examine the effectiveness of each variant in the precision-recall curves, see Fig.\ \ref{fig:precision_recall}.
%The parameter \( \alpha \) serves as a critical cut-off value that distinguishes between similar and dissimilar pairs in our model.
% remove?
As expected, RANSAC clearly outperforms the Super-features curve. Despite both approaches offering competitive performance compared to the baseline models without re-ranking, precision deteriorates rapidly, confirming again challenges in S3LI due to the lack of salient features.

%\subsubsection{Pre-training}
\textbf{Pre-training}: To assess the impact of pre-training (see Sec.\ \ref{subsec:training-pretrain}), we perform an ablation study by initializing the weights with either random values or a pre-trained set. 
Table \ref{tab:robotcar-results-pretraining} shows the important role of pre-training. In our experience, this especially applies for complex models. Also, it reinforces the model's robustness and generalization by reducing its propensity to overfit on the visual modality.

\begin{table}[h!]
\centering
\begin{tabular}{|c|c|c|c|}
\hline
\textbf{Method} & \textbf{Recall@1} & \textbf{Recall@5} & \textbf{Top 1\% recall} \\ \hline
UMF (w/o pre-training) & 70.4 & 81.2 & 85.9 \\ 
\textbf{UMF (w/ pre-training)} & \textbf{73.5} & \textbf{82.9} & \textbf{87.5} \\ \hline
\end{tabular}
\caption{Influence of pre-training on S3LI.}
\label{tab:robotcar-results-pretraining}
\end{table}

%\newpage
\section{Conclusions}
In this paper we have presented the ``Unifying local and global Multimodal Features'' (UMF)
model, a novel place recognition method that fuses local and global features of both visual and LiDAR data using transformers and incorporates re-ranking steps based on single-modality local features. We evaluate our UMF against state-of-the-art baselines on two different domains: urban and planetary. UMF shows superior performance in terms of Recall@$\!N$ in both domains. 
%The effective use of both local and global image features in our model leads to significant performance gains. 
In particular, we observed that our model outperforms significantly previous baselines in the extreme conditions of the S3LI dataset (Mt.\ Etna, Sicily), demonstrating the potential of a sound %\todo{instead: learned? sound?} 
multimodal fusion for place recognition in challenging (unstructured and aliased) scenes.
%Future aims that need to be addressed are the scalability of our approach, as well as resiliency to severely unbalanced information originating from the complementary modalities. 

%\newpage

\balance
\bibliographystyle{IEEEtran}
\bibliography{Bibliografia_TFM}

\end{document}